# RecLight: A Recurrent Neural Network Accelerator with Integrated Silicon Photonics

Febin Sunny, Mahdi Nikdast, and Sudeep Pasricha
Department of Electrical and Computer Engineering
Colorado State University, Fort Collins, CO, USA
{febin.sunny, mahdi.nikdast, sudeep}@colostate.edu

*Abstract*—Recurrent Neural Networks (RNNs) are used in applications that learn dependencies in data sequences, such as speech recognition, human activity recognition, and anomaly detection. In recent years, newer RNN variants, such as GRUs and LSTMs, have been used for implementing these applications. As many of these applications are employed in real-time scenarios, accelerating RNN/LSTM/GRU inference is crucial. In this paper, we propose a novel photonic hardware accelerator called RecLight for accelerating simple RNNs, GRUs, and LSTMs. Simulation results indicate that RecLight achieves 37× lower energy-per-bit and 10% better throughput compared to the state-of-the-art.

## I. INTRODUCTION

Recurrent Neural Networks (RNNs) are a class of Artificial Neural Networks (ANNs) where connections among neurons form a directed graph along a temporal sequence. Such models have internal memory and feedback connections that make them well suited for learning trends and patterns inherent in sequences where the data elements are correlated. As a result, RNNs have been found to perform well for sequence learning tasks, such as speech recognition, human activity recognition, etc. [1]. While recent developments with Transformer models for sequential learning are promising, such models have large parameter counts that are not suited for resource-limited platforms [1].

When learning large sequences of data, simple RNNs [2] face the problem of vanishing gradients, which limits their usability. To alleviate the vanishing-gradients issue, more advanced RNN models have been developed based on Gated Recurrent Units (GRUs) [3] and Long Short-Term Memory (LSTM) [4]. These models are often employed in real-time scenarios, such as in IoT devices with virtual voice assistants and natural language processing abilities. Therefore, there is a critical need for efficiently accelerating such models for edge/IoT environments. However, inference acceleration in RNNs is a challenging task because of the recursive nature of these models and the compute-intensive operations required for large-dimensional sequence data. Moreover, RNNs are very reliant on the activation functions they employ, particularly the *sigmoid* and *tanh* functions. Thus, accelerating RNNs requires unique strategies that differ from those for accelerating other ANN models, such as MLPs and CNNs.

In recent years, several accelerators for RNNs have been proposed [5]-[10]. Most of these efforts aim at accelerating a single RNN variant: LSTMs. However, other RNN models with simple RNNs and GRUs can be useful in resource-constrained scenarios. In particular, GRUs can offer comparable performance as LSTMs while offering faster execution and using less memory. In this paper, we present the design of a novel RNN accelerator called RecLight which can accelerate ANNs that consist of any combination of simple RNNs, GRUs, and LSTMs. Unlike any prior RNN accelerator, we leverage noncoherent integrated silicon photonics. Silicon photonics is already a proven solution for high-throughput communication in the telecom, datacom, and rack-level computing domains, but in recent years it has also shown immense promise to accelerate computations [10]. The use of CMOS-compatible silicon photonic devices and circuits can overcome the energy and performance bottlenecks in conventional electronic accelerators. The novel contributions of this work are as follow:

- The design of a novel noncoherent silicon photonic accelerator targeting accelerating RNN variants;
- A detailed analysis of achievable resolution for RNNs with silicon-photonic microring resonator devices;
- A novel photonic multiply-and-accumulate (MAC) unit design that minimizes power dissipation and energy consumption while maximizing the overall throughput;
- A comprehensive comparison with state-of-the-art electronic RNN accelerators, for sequence learning.

The rest of the paper is organized as follows. Section II presents a background on RNNs and their acceleration with photonic devices. Section III gives an overview of the RecLight architecture. Section IV discusses experimental setup and results, followed by the conclusions in Section V.

## II. BACKGROUND AND RELATED WORK

### A. RNN acceleration

RNN is a term used to denote any ANN model with feedback connections to the neurons in a layer. Such models are used for learning temporal dependencies between elements in a sequence, such as time series data. Due to the simplistic nature of the fundamental block in a simple RNN model, it is prone to exploding/vanishing gradients during training, which prevents the model from learning long-term dependencies in the input data [9]. To learn longer term dependencies, more complex RNN cells such as GRUs and LSTMs can be useful. Compared to simple RNNs, the gates and states used in GRUs and LSTMs make them effective for learning long-term dependencies. The individual cells are typically chained together within a layer, and multiple layers are often stacked together to realize powerful deep RNN models for sequence learning problems.

RNN accelerator-design efforts have mostly focused on LSTM acceleration, possibly owing to the increased popularity of the LSTM models over the other two RNN model variants. The work in [5] presented an FPGA implementation for LSTM acceleration using a software-hardware co-optimization approach. In [6], a similar FPGA implementation approach is used with a compression technique to accelerate LSTM inference. This approach employs block-circulant, instead of sparse matrices, to compress weight matrices. ASIC implementations of LSTM accelerators are proposed in [7] and [8]. In [7], approximate multiplication was employed along with synchronization of the proposed elastic pipeline to maximize the accelerator throughput. The architecture in [8] utilized systolic arrays for acceleration and to reduce memory transfer overhead. Some recent FPGA-based implementations of GRU accelerators have been presented in [9] and [10]. *Unlike these efforts, RecLight supports accelerating all three major RNN variants.*

### B. Silicon photonics for ANN acceleration

Silicon photonics has already been established in literature as energy-efficient, high throughput solution for on-chip communication [11], [12]. Silicon-photonic ANN accelerators have received significant interest in recent years [13]. Optical ANN accelerators can be broadly classified into two types: coherent and noncoherent architectures. <u>Coherent architectures</u> use a single wavelength to operate and imprint weight/activation parameters onto the electrical field amplitude, phase, or polarization of an optical signal [14]. Here, the term *coherent* refers to the physical property of the wave with which it can interfere constructively or destructively on the same wavelength. <u>Noncoherent architectures</u>, such as [15]-[18], use multiple wavelengths, where each wavelength can be used to perform computations in parallel. In these architectures, parameters are imprinted onto the signal amplitude and wavelength-selective devices, such as microring resonators (MRs; see Fig. 1, top left), are used to manipulate individual wavelengths. Existing noncoherent photonic ANN accelerators primarily focus on accelerating CNNs and MLPs (see survey in [13]). *To the best of our knowledge, RecLight is the first RNN accelerator that leverages noncoherent silicon photonics.*

### C. Computations with noncoherent photonic devices

Microring Resonators (MRs) are used as the primary optoelectronic device for computation in noncoherent architectures. As *RecLight* utilizes these devices, we provide a brief background on their operation. An MR is designed to be sensitive to a particular wavelength, called its resonant wavelength ($\lambda_{MR}$), which depends on multiple factors based on:

$$\lambda_{MR} = \frac{2\pi R}{m} n_{eff}, \quad (1)$$

where $R$ is the radius of the MR, $m$ is the order of the resonance, and $n_{eff}$ is the effective refractive index of the device. An MR can modulate (transmit) electronic data over an optical signal $\lambda_{MR}$ with the help of a tuning circuit that can alter $n_{eff}$ in a carefully controlled manner. The MR tuning mechanism can induce an appropriate resonant shift ($\Delta\lambda_{MR}$), to change the output wavelength amplitude (Fig. 1, top left) and realize a scalar multiplication operation. Such tuning is also used to imprint the desired parameters on an optical signal by adjusting an MR's tuning signal (corresponding to the parameter value), and hence varying the signal magnitude through the loss a wavelength experiences as it passes the MR. The tuning mechanism in MRs can be implemented via either microheaters (thermo-optic (TO) tuning [19]) or carrier injection (electro-optic (EO) tuning [20]), thereby inducing a change in $n_{eff}$, which impacts $\lambda_{MR}$, and introduces the appropriate $\Delta\lambda_{MR}$.

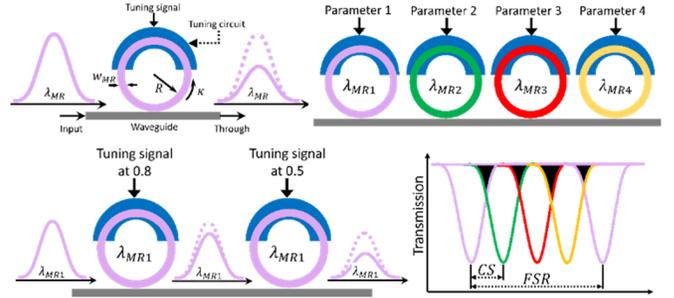

Fig. 1: An MR with tuning circuit (top left) used for tuning wavelengths to reflect parameter values. Such MRs can be placed together to form an MR bank (top right). MRs of the same wavelength can be used to perform multiplication operations (bottom left). The transmission spectrum of an MR bank is shown on bottom right, depicting free-spectral range (*FSR*), channel spacing (*CS*), and inter-channel crosstalk (regions shaded black).

The behavior of a large number of neurons can be emulated in noncoherent architectures by using wavelength-division multiplexing (WDM). To process multiple wavelengths simultaneously, several MRs can be placed together on the same waveguide to form an MR bank (Fig. 1; top right). The number of wavelengths that can be accommodated with WDM depends on the free-spectral range (*FSR*) of the MRs. *FSR* is the spectral distance between two consecutive resonant peaks/modes of the *same* MR. To accommodate a large number of wavelengths, a large *FSR* is required. Moreover, to ensure reliable operation, channel spacing (*CS*), which is the spectral distance between two adjacent (different) MR resonances, must be sufficiently large (see Fig. 1; bottom right). Low *CS* can cause power from adjoining resonances to leak into each other causing inter-channel or heterodyne crosstalk [37] (indicated by the regions shaded black in Fig. 1, bottom right). The next section describes the *RecLight* architecture that addresses these challenges for reliable and high-performance photonic RNN acceleration.

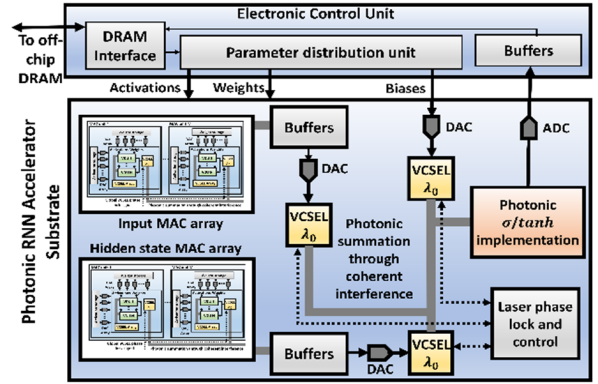

Fig. 2: An overview of the proposed *RecLight* architecture.

### III. RECLIGHT ARCHITECTURE

*RecLight* is a noncoherent photonic architecture that can accelerate inference with simple RNN, GRU, and LSTM based models. An overview of the *RecLight* architecture is shown in Fig. 2. In the following subsections, we describe the *RecLight* architecture and the hardware optimizations we have considered to efficiently accelerate RNNs with *RecLight*.

### A. MR tuning circuit design

*RecLight* makes use of a hybrid tuning circuit where both TO and EO tuning are used to induce $\Delta\lambda_{MR}$. EO tuning is faster (≈ns range) and consumes lower power (≈4 μW/nm), but with a smaller tuning range [20]. In contrast, TO tuning has a larger tunability range, but consumes higher power (≈27 mW/*FSR*) and has higher (≈μs range) latency [19]. The hybrid tuning approach considers the advantages that each tuning mechanism offers while covering for its disadvantages. The feasibility of such a hybrid tuning approach has previously been shown in [21] for silicon photonic devices with low insertion loss. We use this approach for hybrid tuning of MR banks in our architecture. The approach supports efficient operation of MRs with fast EO tuning to quickly induce small $\Delta\lambda_{MR}$ and using the slower TO tuning infrequently for large $\Delta\lambda_{MR}$. To further reduce the power overhead of TO tuning in the hybrid approach, we adapt a method called thermal Eigenmode decomposition (TED), which was first proposed in [22]. Using TED, we can collectively tune all the MRs in an MR bank with lower power consumption. TED also comes with the advantage of alleviating thermal crosstalk noise generated by heat dissipated from adjoining TO circuitries which use microheaters to induce thermal tuning.

### B. MR device design and resolution analysis

By using TED and alleviating thermal crosstalk, which was pointed out to be the main constraint in parameter resolution achievable in noncoherent photonic computation in [23], we can achieve better resolution in *RecLight*. In addition, we consider the inter-channel crosstalk in an MR bank, using the analytical models from [23] (see Fig. 1; bottom right).

As MR count increases, the resulting inter-channel crosstalk prevents good resolution from being achieved, at lower Q-factor values. At sufficiently high Q-factor values (9000 to 10000), even large MR banks can achieve 32-bit resolution, due to the sharper resonance (i.e., higher Q-factor) reducing crosstalk. But high-resolution support comes with the overhead of high-resolution digital-to-analog converters (DACs) and analog-to-digital converters (ADCs), which are power hungry devices. Hence, we consider 16-bit resolution for our parameters, as 16-bit quantized models can achieve comparable performance to full-precision models [24]. Also, the large channel spacing and MR count in banks comes with the need for large *FSR* values, which are difficult to achieve. A larger *FSR* requires smaller radii which introduce higher optical losses in MRs. From our analysis, we found that with *CS* = 2.5 nm and Q-factor = 5000 in MRs, our MR banks can achieve a resolution of 16 bits with up to 15 MRs per bank. Using the models in [25], MR radius (*R*) can be described as:

$$R = \frac{Q\lambda_{MR}\kappa^2}{2\pi^2 n_g\sqrt{1-\kappa^2}}, \qquad (2)$$

where $\kappa$ is the coupling coefficient and $n_g$ is the group index of the MR. We set Q at 5000, for our exploration presented in Fig. 3. For $\lambda_{MR}$ = 1550 nm, waveguide thickness of 220 nm, input waveguide width of 400 nm, and a gap of 100 nm, we performed an exploration for *R* and MR waveguide width ($w_{MR}$) while satisfying the Q-factor requirement of 5000. To obtain the corresponding $\kappa$ and $n_g$ values, we performed detailed device-level simulations with the ANSYS Lumerical tool [26].

The results from this experiment are presented in Fig. 3. To avoid strong higher-order mode excitation when increasing $w_{MR}$, we selected $w_{MR}$ and *R* to be 700 nm and 5 μm (green circle in Fig. 3), respectively. Note that a smaller *R* will impose higher optical losses. The resulting FSR is 19.3 nm, which is sufficient for achieving our 16-bit parameter resolution goal in *RecLight*.

The parameters obtained from this analysis are used to guide our architectural analysis, presented in Section IV.

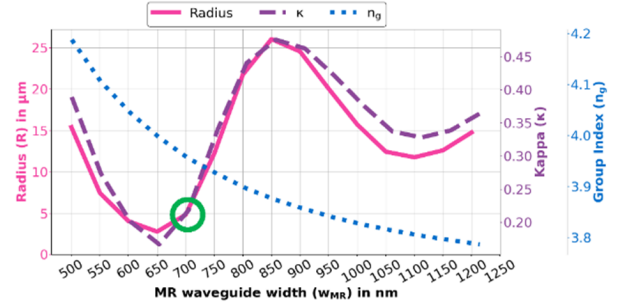

Fig. 3: MR design exploration with the selected MR design (R=5 μm) highlighted by the green circle.

### C. VDU and MAC unit design

Effective ANN inference acceleration requires accelerating the most time-consuming operations during inference, which happen to be matrix-multiplication operations. This also holds true for RNNs, as most operations in RNN cells involve multiplication between matrices (of weights, inputs, etc.). These operations can be decomposed into vector-dot-product operations, as discussed for CNNs in [23]. The Vector-Dot-product Units (VDUs) in *RecLight,* as shown in Fig. 4(a), are photonic computation units designed to perform vector-dot-product operations. RNN weights and activations are routed to individual MR tuning circuits using 16-bit DACs (to support 16-bit parameter resolution). To reduce the power consumption in the DACs, which can be substantial, we use a local parameter storage mechanism within the VDU that relies on memristors. A memristor cell is integrated into the EO tuning mechanism of an MR (see Fig. 4(a)). The conductance of the memristor alters the biasing voltage being applied across the EO tuning junction in the MR. This conductance can in turn be tuned with an appropriate signal from the DAC. As the memristor can hold this conductance value once the voltage across it is removed, we can use the same DAC array to tune multiple MR banks. For this, we consider splitting the MR banks in a VDU across multiple waveguides ($N_{WG}$). If the VDU handles a vector granularity of *v*, this split allows us to use only $2v/N_{WG}$ DACs instead of the initial 2*v* DACs required. While this approach does incur some penalty in the form of slightly increased latency, the power benefits it brings far outweighs this penalty. The stored conductance in a memristor cell allows EO tuning to leverage the stored parameter to set the junction voltage across the tuning junction in the MR. Banks of such MRs within the VDUs perform the dot-product operations within the RNN cells mapped to them. These banks can also be tasked with accelerating fully connected (FC) layers which usually come after the RNN layers in deep RNN models used in many sequence-learning applications. To support both positive and negative values of parameters involved, we use separate positive and negative parameter arms in a VDU, for the same waveguide. The sum obtained from the negative arm is subtracted from the sum from the positive arm using a balanced photodetector (PD) arrangement, shown as BPD in Fig. 4(a).

Multiple VDUs are combined to form a photonic Multiply and Accumulate (MAC) unit as shown in Fig. 4(b). The VDUs in a MAC unit share the laser source and the DAC array between them. The laser sources we use in *RecLight* are vertical cavity surface-emission laser (VCSEL) [27], [38] arrays. The shared VCSEL array allows for reusing the same wavelengths across multiple ($N$) VDUs, thereby reducing the VCSEL requirement and laser power consumption. This VCSEL-reuse also allows our architecture to attain the large channel spacing requirement (see Section III.B) to attain 16-bit resolution. Splitting the MR banks across $N_{WG}$ waveguides also helps further reduce the laser power consumption and possible inter-channel crosstalk. This split does incur a splitter loss (considered in our analysis), but the advantages it brings in terms of power consumption and robustness in operation are considerable.

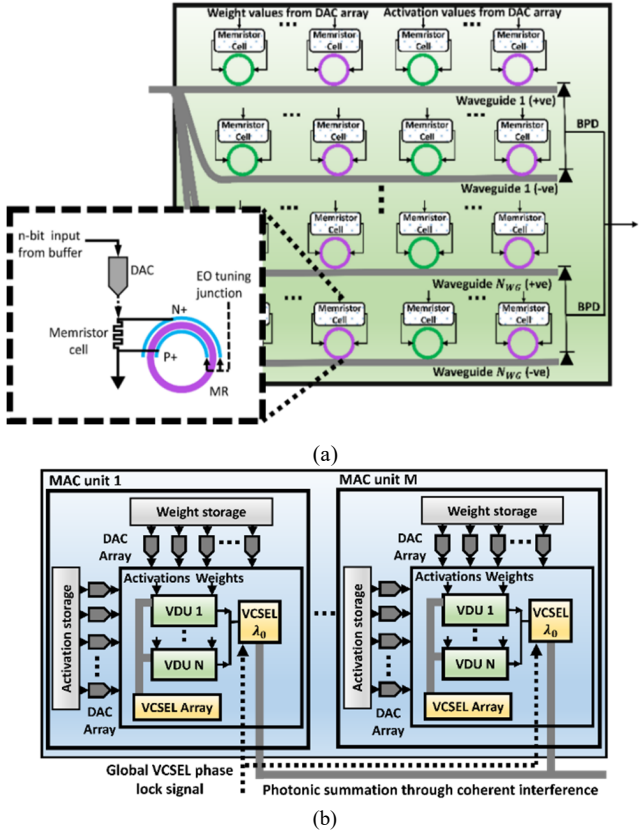

(a)

(b)

Fig. 4: (a) VDU showing an MR bank with memristor cells for local parameter storage (BPD: Balanced photodetector). Inset: EO tuning control for memristor cell in VDU. (b) A MAC array comprised of $M$ MAC units, each with $N$ VDUs. Each MAC unit has a vertical cavity surface-emission laser array (VCSEL array) driven using the output from the VDU array.

To combine the partial sums generated by the MAC units, we employ coherent photonic summation. For this, we use an electrical signal from the VDU array to drive a VCSEL. Across MAC units, these driven VCSELs all generate the same wavelength $\lambda_0$ that, when introduced into the same waveguide, undergoes interference to generate the sum from a MAC unit array. To ensure coherent summation, we use a laser phase locking mechanism [28]. It ensures that VCSELs' output signals are in phase and hence constructive interference can occur. The output from the MAC unit array is added to the corresponding bias value optically, depending on which gate matrices were deployed. The bias value is fed directly to a $\lambda_0$ VCSEL, through a 16-bit DAC, for driving it, and photonic coherent summation is performed to obtain the summed output.

### D. Implementation of the non-linear unit

RNN cells require specific non-linear activation functions (*sigmoid* and *tanh*). While most photonic ANN accelerators assume that activation functions are implemented electronically [10], this can lead to high overhead due to frequent opto-electronic conversions that would be needed for each RNN cell. To reduce such overhead, we consider an optoelectronic implementation of the activation functions. The work in [29] implemented non-linear functions such as *sigmoid ($\sigma$)* using silicon photonic components (see Fig. 5). In [27], a photonic control unit is used to drive $i+$ and $i-$ signals that are fed to the EO tuning circuitry of the MR. $I_b$ and $I_h$ are applied to, respectively, the EO tuning and TO tuning in the MR. But in our architecture, the required saw-tooth waveform signals can be generated by a more efficient electronic circuit as we only need to generate $\sigma$. Note that *tanh* can be also implemented based on $\sigma$ (for input signal $x$) as:

$$tanh(x) = 2\sigma(2x) - 1. \quad (3)$$

To implement these activation functions, we use two Semiconductor-Optical Amplifiers (SOAs) [36], each providing a 100% gain to the input signal (Fig. 5). The stored result is fed to a power gated electronic subtractor circuit to obtain the *tanh* value. The circuit in Fig. 5 can be reconfigured to implement both $\sigma$ and *tanh*, as enabling SOAs and the subtractor circuit will generate the *tanh* function and $\sigma$ otherwise.

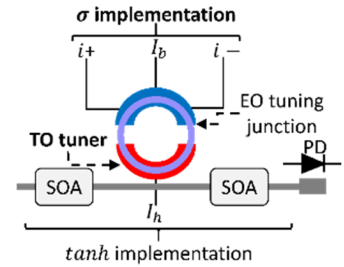

Fig. 5: *Sigmoid ($\sigma$)* [27] and *tanh* implementation for *RecLight*.

### E. RecLight architecture

As shown in Fig. 2, the architecture of *RecLight* is designed to accelerate all the three RNN variants, including simple RNNs, GRUs, and LSTMs. Each VDU (see Fig. 4(a)) in the architecture is assigned vectors with vector granularity of $v$ to operate on. $N$ VDUs along with their respective shared VCSEL array and BPDs, and $\lambda_0$ VCSELs (for summation) form a single MAC unit (see Fig. 4(b)). Each MAC unit has its own local weight and activation parameter storage and associated DAC array. Each DAC array holds $v$ 16-bit DACs to feed the parameters to one VDU at a time. $M$ MAC units form a MAC array. Each MAC array is tasked with an RNN cell gate-level matrix multiplication. Each type of RNN is composed of temporal iterations of fundamental cells, each of which has gates associated with it. To accelerate an RNN, this fundamental cell operation and the associated gate operation must be accelerated. Our MAC units are designed to take into account the sequential nature in which the gate level RNN operations are performed.

Specifically, each gate operation requires an input state MAC operation and a hidden state MAC operation. Our architecture has MAC arrays specifically assigned for input and hidden state MAC-operation acceleration. In an RNN cell, two weight matrices, i.e., the hidden state vector and the input vector, along with the corresponding gate's bias vector are involved in a gate-level operation. To reflect this, two MAC arrays, each handling one of the two matrices, are designed with $M$ MAC units each. The outputs from the two MAC arrays are photonically summed, to which the bias parameter can be added photonically without any electrical-to-optical conversion. The coherent photonic summation also allows us to subject the overall sum to the photonic non-linearity implementation. The non-linearity being used depends on the gate being operated on (Section III.D). The result is collected in a storage unit where minor post-processing is performed if needed. In this manner, layers of RNNs can be processed in *RecLight*. Moreover, fully connected (FC) layers (found in some deep RNN models) can also be accelerated by decomposing and mapping them to the VDUs in the architecture.

Table I: RNN models considered for analysis.

| Weather data time series prediction | | | |
|---|---|---|---|
| Model | Total parameters | MAE (32-bit) | MAE (*RecLight*) |
| RNN | 152,976 | 0.4820 | 0.489 |
| GRU | 170,880 | 0.5782 | 0.5844 |
| LSTM | 217,696 | 0.5621 | 0.5650 |
| IMDB sentiment analysis | | | |
| Model | Total parameters | Accuracy (32-bit) | Accuracy (*RecLight*) |
| RNN | 2,216,137 | 73.8% | 72.75% |
| GRU | 2,691,713 | 75.3% | 74.7% |
| LSTM | 3,156,236 | 77.3% | 76.8% |
| PTB dataset for language modelling | | | |
| Model | Total parameters | Perplexity (32-bit) | Perplexity (*RecLight*) |
| RNN | 11,015,000 | 131.45 | 131.63 |
| GRU | 13,952,000 | 97.7 | 98.5 |
| LSTM | 14,615,000 | 66.02 | 65.78 |

## IV. EXPERIMENTS AND RESULTS

To evaluate the effectiveness of *RecLight*, we performed several simulation-based analyses. We consider three datasets to build RNN models: a time series analysis based on the weather dataset from [30], the IMDB sentiment analysis dataset, and the Penn Treebank (PTB) dataset for language modeling. We designed an RNN, GRU, and LSTM based ANN model each for these datasets, details of which are provided in Table I.

Table II: Parameters considered for analysis of *RecLight*.

| Devices | Latency | Power |
|---|---|---|
| EO Tuning [20] | 20 ns | 4 $\mu$W/nm |
| TO Tuning [19] | 4 $\mu$s | 27.5 mW/*FSR* |
| VCSEL [27] | 0.07 ns | 1.3 mW |
| Photodetector [32] | 5.8 ps | 2.8 mW |
| DAC (16 bit) [33] | 0.33 ns | 40 mW |
| ADC (16 bit) [34] | 14 ns | 62 mW |
| Memristor cell [35] | 0.1 ns | 0.07 $\mu$W |

We designed a *RecLight* simulator in Python to estimate performance and energy costs, by modeling the microarchitecture of the MAC units as described in Section III.C. The simulator performs layer-wise decomposition of RNN parameters into vectors, mapping them onto the modeled MAC units, and analyzes latency and energy consumption for the mapped operations. We parameterized the energy and latency requirements of the devices, as per the parameters presented in Table II, which are based on fabricated silicon photonic devices. We used Tensorflow 2.3 with Qkeras [31] for analyzing model accuracy across different parameter resolutions. From our analysis, the 16-bit quantized RNN models, as they are deployed in our architecture, perform with comparable accuracies to models with full precision (32-bit) parameters, as can be seen from Table I. Table II shows the optoelectronic parameters considered for the simulation-based analysis with *RecLight*. As discussed in Section III.E, *RecLight* design involves parameters $v$ (vector granularity), $N$ (number of VDUs per MAC unit), $M$ (number of MAC units), and $N_{WG}$ (number of waveguides in a VDU). We performed an analysis to determine the best [$v$, $N$, $M$, $N_{WG}$] configuration possible for *RecLight* in terms of throughput (giga-operations-per-second (GOPS)) and energy-efficiency (energy-per-bit (EPB)). The result of this exploration is presented in a scatterplot in Fig. 6. From this exploration, we can identify the *RecLight* architecture configuration with the best EPB/GOPS ratio, across all the models considered, with the configuration [15, 15, 40, 10] shown by the pink star in Fig. 6. This *RecLight* configuration is used for further analyses.

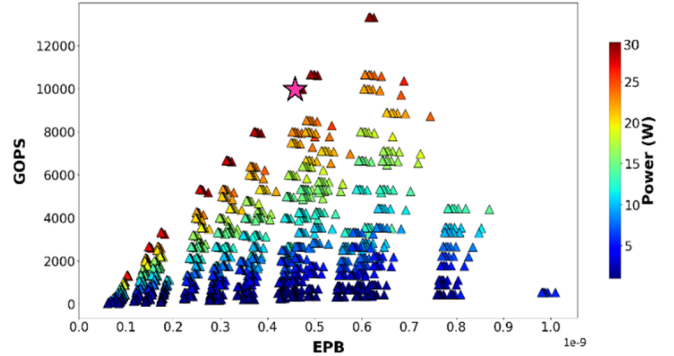

Fig. 6: Architectural exploration analysis for *RecLight*, with the aim to find the optimal [v, N, M, $N_{WG}$] configuration with the best energy-efficiency and throughput. The best configuration, which is [15, 15, 40, 10], has the lowest EPB/GOPS value and is indicated using a pink star

### A. Comparison to state-of-the-art RNN accelerators

To analyze how *RecLight* compares to other accelerators when executing RNN models, we compare it against state-of-the-art electronic RNN accelerators: BBSL [5], C-LSTM [6], ELSA [7], and Chipmunk [8], which are LSTM accelerators, and with DeltaRNN [9] and EdgeDRNN [10], which are GRU accelerators. We do not show comparison results with other photonic accelerators as there is no prior work on *noncoherent* photonic RNN accelerators. We used energy and performance information as reported in the selected accelerators in our analysis to estimate the EPB and GOPS metrics for each accelerator, when executing the models described in Table 1.

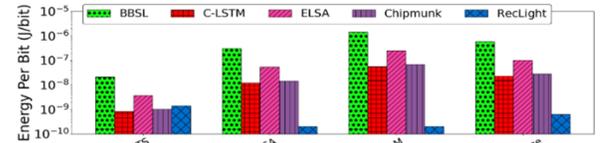

Fig. 7: EPB comparison between LSTM acceleration. TS = time series, SA=Sentiment analysis, and LM= language modeling.

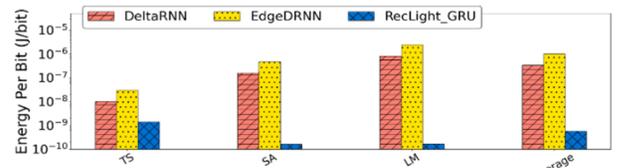

Fig. 8: EPB comparison between GRU acceleration. TS = time series, SA =Sentiment analysis, and LM= language modeling.

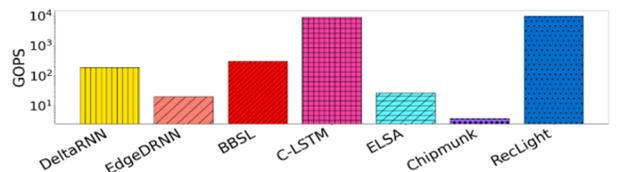

Fig. 9: Throughput comparison among accelerators.

Fig. 7 illustrates an energy-per-bit (EPB) comparison between the *RecLight* and the LSTM accelerators considered.

We have not considered simple RNN and GRU model acceleration on the four accelerators from prior work as they are not designed to support these models. From the results, *RecLight* shows much lower EPB for LSTM acceleration. This is in part because of the low power consumption our accelerator achieves due to our device, circuit, and architecture level optimizations discussed in Section III, and due to the low latency operation of the photonic substrate. *RecLight* does show higher EPB for the time series (TS) LSTM model as the model is simpler (see Table I) and does not allow amortizing the static power overhead in our architecture. On average, *RecLight* obtains 956×, 37×, 167×, and 45× lower EPB than BBSL, C-LSTM, ELSA, and Chipmunk accelerators, respectively.

Fig. 8 shows an EPB comparison between the GRU accelerators DeltaRNN [9] and EdgeDRNN [10], and *RecLight* running GRU models for inference (see Table 1). An EPB trend similar to what is shown in Fig. 7 can be observed here again for *RecLight*, for the same reasons discussed earlier. From our analysis, *RecLight* obtains 1730× and 570× better EPB than DeltaRNN and EdgeDRNN accelerators, respectively.

Finally, Fig. 9 shows the GOPS comparison across all the accelerators. *RecLight* achieves 51.9×, 494.25×, 33.3×, 1.1×, 370.4×, and 2631.6× better throughput (y-axis is in log scale) in terms of GOPS compared to the DeltaRNN, EdgeDRNN, BBSL, C-LSTM, ELSA, and Chipmunk, respectively. The higher GOPS with *RecLight* can be attributed to its high-speed photonic computation with very few intermediate optical-to-electrical conversions.

## V. Conclusions

In this paper, we presented the first noncoherent photonic accelerator for RNN models, called *RecLight*. Our accelerator exhibits energy-per-bit improvements that range from 37× to 1730× when compared with six state-of-the-art electronic RNN accelerators. *RecLight* also demonstrates up to 2631.6× better throughput than these electronic RNN accelerators. These results demonstrate the promising low-energy and high-throughput inference acceleration capabilities of our *RecLight* architecture. While in this work we focused entirely on the optoelectronic hardware design of our accelerator, with better software techniques for compressing RNN models, even better throughput and energy-efficiency improvements might be achievable with silicon-photonic-based accelerators.

## Acknowledgment

This work was supported by National Science Foundation (NSF), through grants CCF-1813370 and CCF-2006788.